\documentclass{article} 
\usepackage{iclr2027_conference,times}

\usepackage{amsmath,amsfonts,bm}

\def\eqref#1{equation~\ref{#1}}

\def\1{\bm{1}}

\DeclareMathAlphabet{\mathsfit}{\encodingdefault}{\sfdefault}{m}{sl}
\SetMathAlphabet{\mathsfit}{bold}{\encodingdefault}{\sfdefault}{bx}{n}

\usepackage{url}

\usepackage{graphicx}
\usepackage{booktabs}
\usepackage{xcolor}
\usepackage{enumitem}
\usepackage{listings}
\usepackage[colorlinks=true,linkcolor=blue,citecolor=blue,urlcolor=blue]{hyperref}

\title{Lightweight Reasoning Transfer via \\ Chain-of-Thought Vector Projection}

\author{
  Yulin Qi, Shuo Xu, Shuxia Lin, Qingjian Ni, Xu Yang \& Xin Geng \\
  School of Computer Science and Engineering, Southeast University
}

\iclrfinalcopy 

\begin{document}
\maketitle
\fancyhead{}
\renewcommand{\headrulewidth}{0pt}

\begin{abstract}
Transferring reasoning capabilities from large language models to smaller ones often requires extensive teacher supervision or costly student adaptation. We study projected cross-model Chain-of-Thought (CoT) Vector transfer, a two-stage approach that keeps both models frozen. First, we learn a task-level CoT Vector in the teacher using dataset-provided solution traces and final-answer supervision. We then train a lightweight linear projector to map the fixed vector into a smaller student’s residual space. Student-side training uses only final-answer cross-entropy on a small subset of task training examples; it requires neither teacher outputs nor intermediate solution traces. We select the student injection layer and intervention scale on a labeled support set and report results on the disjoint remainder of each evaluation split. On GSM8K and MATH, the projected vector improves base students by 10.37–21.98 accuracy points across the evaluated teacher–student settings. On four code-generation benchmarks, the Qwen2.5-7B-Instruct student gains 3.23–4.67 pass@1 points, while the base Qwen2.5-7B student gains up to 9.65 points. These results show that a teacher-side reasoning vector can be reused across smaller students with different residual dimensions through projector-only adaptation.
\end{abstract}

\section{Introduction}
\label{sec:introduction}

Large language models (LLMs) have achieved impressive performance on complex reasoning tasks, particularly as model scale and training compute increase. Chain-of-thought (CoT) prompting further unlocks multi-step reasoning abilities that are often most pronounced in sufficiently large models~\citep{wei2022chain}. Nevertheless, the substantial memory and computational requirements of large models complicate deployment in resource-constrained, privacy-sensitive, or latency-critical applications~\citep{dettmers2022llmint8}. Smaller language models therefore remain indispensable, motivating a fundamental question: \emph{how can the reasoning capability of a strong large model be transferred to a substantially smaller model without incurring an equally expensive adaptation process?}

\begin{figure}[t]
    \centering
    \includegraphics[width=\linewidth]{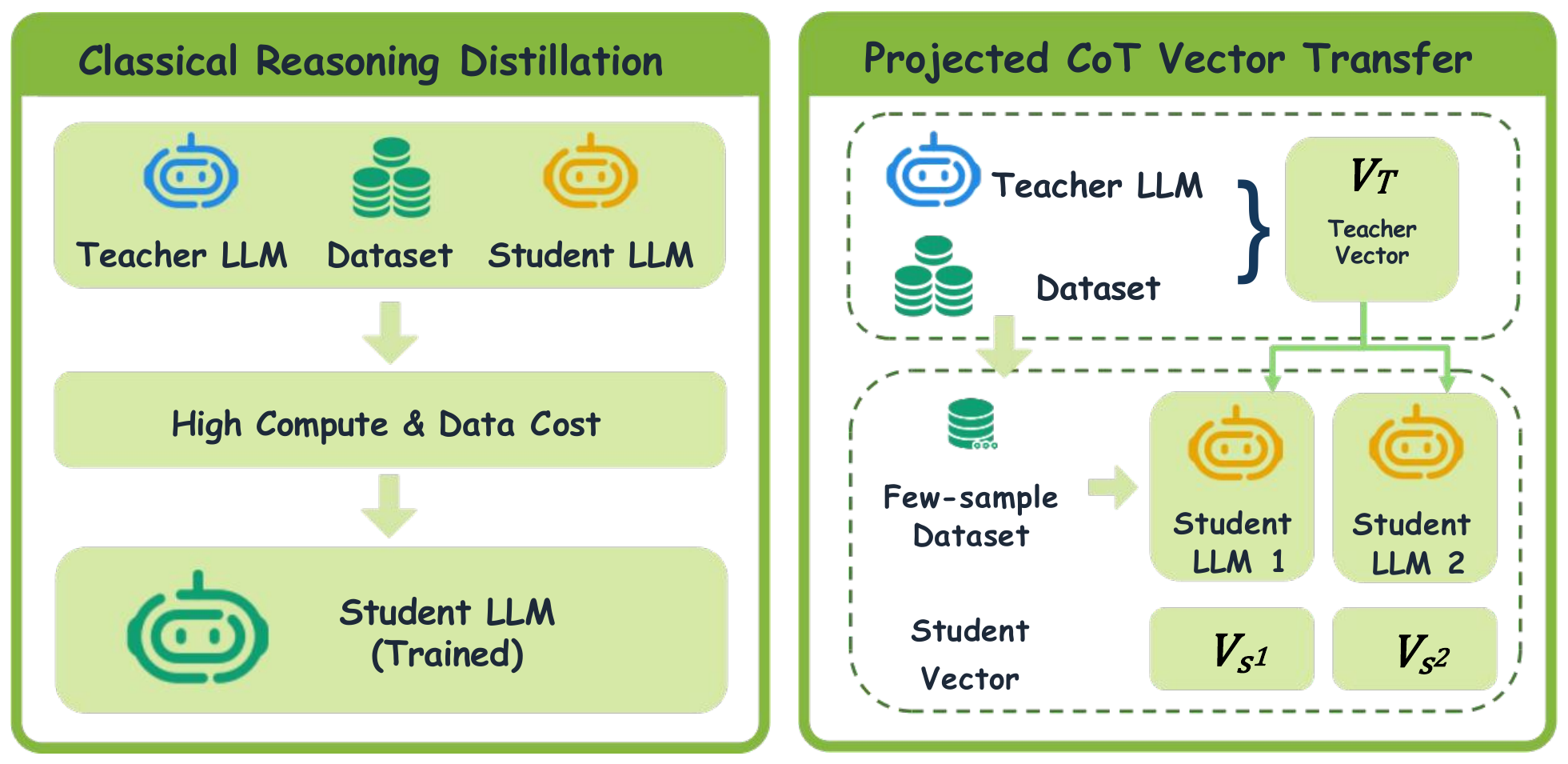}
    \caption{Schematic comparison of classical knowledge distillation and
    projected CoT Vector transfer. The teacher-side vector can be learned
    once and reused for multiple frozen students, each adapted with a
    small training subset.}
    \label{fig:distillation-comparison}
\end{figure}

Knowledge distillation is a widely used approach for transferring capabilities from a teacher model to a student model. Classical distillation trains the student to match the teacher's output distribution~\citep{hinton2015distilling}, while later methods additionally align intermediate representations~\citep{jiao2020tinybert}. For reasoning tasks, rationale-based distillation uses teacher-generated explanations or CoT traces as additional supervision~\citep{ho2023reasoning,hsieh2023distilling}. Although effective, these approaches commonly require generating and storing substantial teacher supervision, training a large portion of the student model, or optimizing multiple objectives for outputs, rationales, and hidden states. Thus, while knowledge distillation reduces the deployment cost of the resulting student, the transfer process itself can remain expensive. This motivates the development of a lightweight transfer mechanism that avoids full-model adaptation and repeated use of teacher-generated CoT traces as student-side supervision. Figure~\ref{fig:distillation-comparison} contrasts conventional student training with our two-stage transfer of a reusable teacher-side vector.

Recent studies on \emph{task vectors} suggest a different perspective on how model capabilities can be represented. Rather than representing a task only through model parameters or natural-language demonstrations, these studies show that task-relevant information can be localized in compact activation-space representations. \citet{hendel2023incontext} demonstrate that in-context demonstrations can be compressed into a single task vector that recovers task behavior, while \citet{todd2024function} identify function vectors that causally induce learned input--output mappings. Building on this activation-based paradigm, \citet{li2026cotvectors} introduce CoT Vectors, compact representations designed to encode task-general, multi-step reasoning knowledge. They propose both activation-difference-based Extracted CoT Vectors and gradient-optimized Learnable CoT Vectors, showing that these vectors can guide reasoning while the underlying language model remains frozen. They also test direct cross-model injection between two same-width Qwen2.5-Math-7B variants by adding a source-model vector to the target model's activations. These findings motivate a natural possibility: if a model capability can be embedded into a single activation vector, then the vector may serve as a lightweight carrier of that capability across models.

Transferring such a vector to a smaller student with a different residual dimension, however, remains challenging: direct vector injection is ill-defined for such teacher--student pairs. This raises our central question: \emph{can a CoT Vector learned in a strong teacher be transformed for use in the residual stream of a smaller student with a different residual dimension while preserving its usefulness for reasoning?}

In this work, we investigate this question on mathematical reasoning and code generation benchmarks, which require multi-step problem solving and permit objective evaluation. We introduce a two-stage framework comprising \emph{teacher-side CoT Vector learning} and \emph{student-side transfer}. During teacher-side CoT Vector learning, we adopt the Learnable CoT Vector objective of \citet{li2026cotvectors} and keep the teacher model frozen. Each training example is processed by the teacher along two paths. The injected path contains the prompt and final answer but omits the intermediate solution sequence, and the learnable vector is injected into the teacher. The reference path contains the same prompt together with the dataset-provided reference solution sequence, including the final answer. We learn the vector using final-answer cross-entropy on the injected path and a representation-level KL alignment between the two paths. Repeating this procedure over the training set produces a task-level vector for each dataset and candidate block, shared by all examples under that configuration. During student-side transfer, a lightweight projector converts this fixed vector into a student-compatible residual vector for injection into the student. Both backbone models remain frozen throughout the framework. Only the projector is trained on a subset of the task training data using final-answer cross-entropy, without accessing teacher logits or hidden states, aligning teacher--student representations, or using intermediate solution-sequence tokens. We select the injection configuration on a small labeled subset of each benchmark's official evaluation split and report performance on the remaining problems.

Across GSM8K, MATH, and four code-generation benchmarks, the projected CoT Vector improves smaller frozen students in many evaluated settings. These experiments use same-family teacher--student pairs that differ in model scale and residual dimension, showing that direct dimensional compatibility is not required for projected transfer. Student-side transfer trains only a projector on a small subset of task training examples. Because teacher-side vector learning is independent of the student, the same vector can be reused across different students by training a separate projector for each, without further access to the teacher. These results suggest that a teacher-side CoT Vector can provide task-level guidance that a student can use with limited student-side training data.

Our contributions are summarized as follows:
\begin{itemize}
    \item We formulate \emph{projected cross-model CoT Vector transfer} for teacher--student pairs with different scales and incompatible residual dimensions, extending prior direct transfer between same-width models.
    \item We develop a two-stage framework with separate teacher-side vector learning and student-side transfer: the teacher vector is learned independently of any particular student, and each student adapts the fixed vector through a lightweight projector without further access to the teacher. Both backbones remain frozen.
    \item We evaluate the method on mathematical reasoning and code generation benchmarks, finding gains in many settings with projector training on a small subset of task examples using final-answer supervision.
\end{itemize}

\section{Related Work}
\label{sec:related_work}

\paragraph{Reasoning Distillation.}
Knowledge distillation transfers predictive distributions or intermediate features from a teacher to a student \citep{hinton2015distilling,jiao2020tinybert}. For reasoning tasks, rationale-based methods often train smaller models with teacher-generated explanations or CoT traces \citep{magister2023teaching,ho2023reasoning,hsieh2023distilling}. SCOTT further targets the consistency and faithfulness of distilled rationales \citep{wang2023scott}. These approaches build on evidence that explicit intermediate steps can improve reasoning performance \citep{wei2022chain}.
Recent long-chain reasoning distillation methods further tailor textual supervision to the student's capacity. P-ALIGN uses adaptive prefix alignment to construct suitable reasoning supervision~\citep{liu2026palign}, while ORION refines teacher CoT traces through error-aware self-reflection informed by the student's reasoning errors~\citep{wu2026orion}.
Rationale-based reasoning distillation, however, communicates the teacher's reasoning primarily through instance-specific textual traces. We instead consider a different transfer unit: a fixed, task-level reasoning representation learned from the teacher.

\paragraph{Task Vectors and Activation Steering.}
Transformer activations can encode compact representations of tasks and behaviors. Task and function vectors derived from hidden states capture information induced by in-context demonstrations and can causally elicit corresponding input--output behavior \citep{hendel2023incontext,todd2024function}. These activation-space representations differ from weight-space task vectors, which are parameter differences between pretrained and task-finetuned models and can be composed through task arithmetic \citep{ilharco2023editing}. Activation-steering methods more broadly modify intermediate representations at inference time to control model behavior without parameter updates \citep{zou2023representation,turner2023activation,rimsky2024steering}. Within this paradigm, \citet{li2026cotvectors} introduce Extracted and Learnable CoT Vectors for multi-step reasoning and demonstrate preliminary direct cross-model transfer from Qwen2.5-Math-7B-Instruct to Qwen2.5-Math-7B, whose matching residual dimensions permit direct vector reuse. Another line of work compresses reasoning traces into latent representations: \citet{chang2026struggle} render textual CoT traces as images and use their visual features to construct a discrete latent vocabulary for autoregressive reasoning. This represents a different granularity of reasoning information from a fixed, task-level CoT Vector. Building on the Learnable CoT Vector formulation, we instead study projected transfer of a single fixed, task-level vector to smaller students with incompatible residual dimensions, for which direct vector reuse is ill-defined.

\paragraph{Cross-Model Representation and Activation Transfer.}
Representation-based distillation aligns intermediate teacher and student features, often through learned projections \citep{romero2015fitnets,jiao2020tinybert}, while model stitching learns mappings between networks primarily to compare or compose their representations \citep{bansal2021revisiting}. Cross-model transfer has also been studied for continuous prompts, using projectors or relative representation spaces to preserve task semantics across language models \citep{su2022transferability,wu2024zeroshot}. More directly, recent studies have explored cross-model activation mappings for concept and safety interventions \citep{huang2025crossmodel,oozeer2025activation}. \citet{chen2025linear} transfer steering vectors, probes, and sparse-autoencoder weights through affine residual-stream mappings, while \citet{bello2025linear} study affine representation transfer for steering across model scales. The prompt-transfer and activation-mapping studies above establish the feasibility of transferring continuous representations and activation-space interventions across models. However, they primarily concern prompt embeddings, concept or behavior directions, and general linear features. Our setting differs in both the transferred representation and the adaptation supervision: we transfer a fixed teacher-side CoT Vector to a smaller student through a lightweight projector. During student-side transfer, the projector is trained solely with the student's final-answer cross-entropy, without accessing teacher logits or aligning teacher--student hidden states.

\section{Method}
\label{sec:method}

\subsection{Problem Setup and Overview}
\label{sec:problem}
\label{sec:overview}

Let $T$ and $S$ denote teacher and student models with residual
dimensions $d_T$ and $d_S$, respectively. Both backbones remain frozen
throughout our two-stage framework. First, we learn a task-level CoT
Vector $\bm{v}_T\in\mathbb{R}^{d_T}$ in the teacher using
dataset-provided solution traces and final-answer supervision,
independently of the student (Section~\ref{sec:teacher-vectors}). We
then freeze $\bm{v}_T$ and train a lightweight bias-free linear
projector to map it into the student's residual space, using only
final-answer supervision on a small subset of task training examples
(Section~\ref{sec:transfer}).

\begin{figure}[t]
    \centering
    \includegraphics[width=\linewidth]{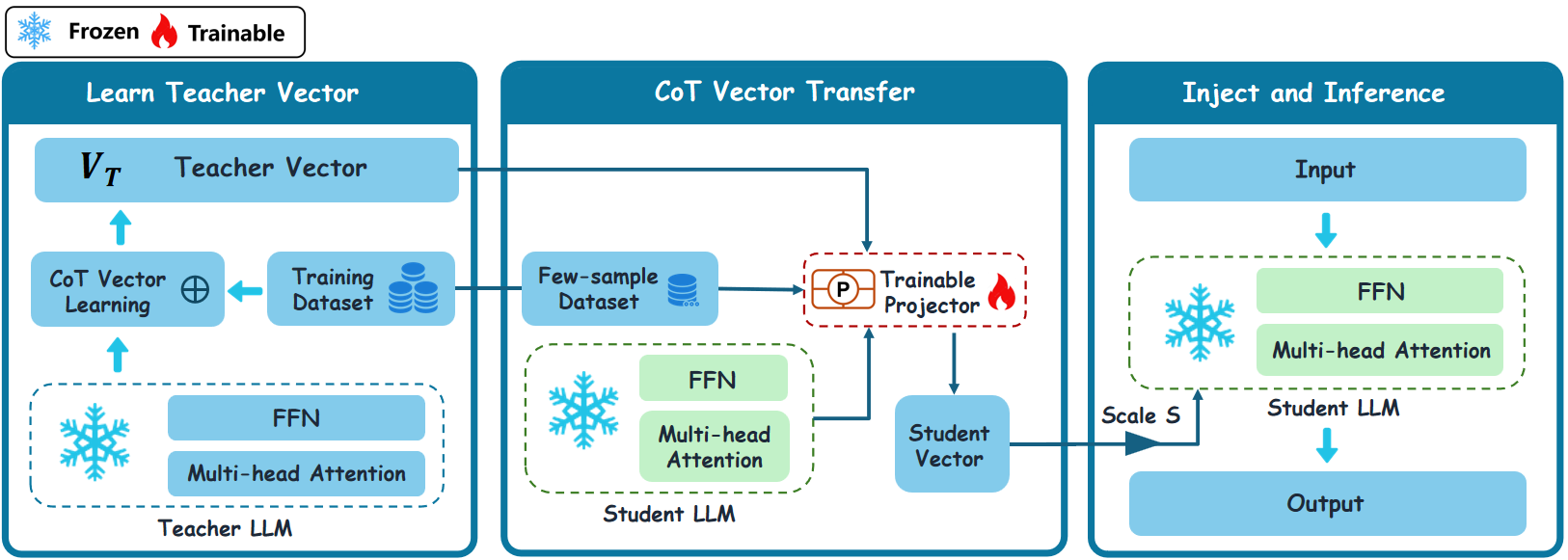}
    \caption{Overview of teacher-side CoT Vector learning, student-side
    projection, and injection into a selected student block. The teacher
    and student backbones are frozen; the student-side projector is trained
    using a small subset of task data. The student vector $V_s$ shown
    corresponds to $\bm{z}=P\bm{v}_T$ before applying scale $s$.}
    \label{fig:method-pipeline}
\end{figure}

We select the student injection block and inference scale on a labeled
support set before inference (Section~\ref{sec:configuration}).
Figure~\ref{fig:method-pipeline} illustrates teacher-side vector learning,
student-side projection, and inference-time intervention.

For a ground-truth final answer, let $\bm{y}_T$ and
$\bm{y}_S$ be its teacher- and student-tokenized sequences,
with respective lengths $N_T$ and $N_S$.

\subsection{Teacher-Side CoT Vector Learning}
\label{sec:teacher-vectors}

Following \citet{li2026cotvectors}, we learn a separate task-level CoT
Vector $\bm{v}_T^{(L)}\in\mathbb{R}^{d_T}$ at a teacher block $L$,
while keeping the teacher frozen. For each training example, the same
prompt is processed along two paths. The reference path contains the prompt
and the dataset-provided reference solution sequence, including the final
answer. The injected path contains only the prompt and the final answer. On
the injected path only, $\bm{v}_T^{(L)}$ is added to the output of decoder
block $L$ at every sequence position; the reference path is left
unintervened.

To align the two paths, we compare their final-normalization-layer
representations at completion positions. The reference completion contains
the dataset-provided solution and final answer; the injected completion
contains only the final answer. Let $N_{\mathrm{ref}}$ be the number of
unmasked tokens in the reference completion. We select the last
$K=\min(N_{\mathrm{ref}},N_T)$ unmasked completion positions from each path.
Thus, $K$ is the number of position pairs used for alignment; we pair the
selected positions in their original left-to-right order. This positional
pairing does not require the tokens in the two paths to match. For
$i=1,\ldots,K$, let
$\bm{h}_i^{\mathrm{ref}}\in\mathbb{R}^{d_T}$ and
$\bm{h}_i^{\mathrm{inj}}\in\mathbb{R}^{d_T}$ denote the outputs of the
teacher's final normalization layer at the $i$-th retained position of the
reference and injected paths, respectively. We treat the coordinates of these
representations as logits and apply a softmax over the residual dimension. The
teacher-vector objective is
\begin{equation}
  \mathcal{L}_{\mathrm{vec}}^{(L)} =
  \frac{\tau^2}{K}
  \sum_{i=1}^{K}
  D_{\mathrm{KL}}\!\big(q_i^{\mathrm{ref}} \,\|\,
  q_i^{\mathrm{inj}}\big)
  + \lambda_{\mathrm{CE}}
  \mathrm{CE}\big(\bm{o}^{\mathrm{inj}}_{\mathrm{ans}},\ \bm{y}_T\big),
  \label{eq:teacher-vector}
\end{equation}
where
$q_i^{\mathrm{ref}}=\operatorname{softmax}(\bm{h}_i^{\mathrm{ref}}/\tau)$
and
$q_i^{\mathrm{inj}}=\operatorname{softmax}(\bm{h}_i^{\mathrm{inj}}/\tau)$,
with $\tau{=}1$; the $\tau^2$ prefactor is retained to match the original
formulation. The KL term aligns the reference and injected-path
representations over residual coordinates, with the reference distribution
as its first argument. Here $\bm{o}^{\mathrm{inj}}_{\mathrm{ans}}$ denotes
the teacher's vocabulary logits at the final-answer positions of the
injected path; the CE term is the mean token-level negative log-likelihood
of the teacher-tokenized ground-truth answer $\bm{y}_T$. We use
$\lambda_{\mathrm{CE}}{=}0.5$. We learn one vector for each dataset and
candidate teacher block $L$, suppressing the dataset index in the notation for readability. After training, $\bm{v}_T^{(L)}$ is frozen and can be reused
across student-side projectors.

\subsection{Student-Side Transfer}
\label{sec:transfer}

Given a fixed teacher vector $\bm{v}_T$, we consider one candidate
student block at a time and train a bias-free linear projector $P$
for that block. The
projected vector is
\begin{equation}
  \bm{z}=P\bm{v}_T,
  \qquad P\in\mathbb{R}^{d_S\times d_T}.
  \label{eq:projection}
\end{equation}
Let $\bm{H}\in\mathbb{R}^{n\times d_S}$ denote the output residual
stream of the chosen block for a sequence of length $n$. We add the
projected vector at each sequence position:
\begin{equation}
  \bm{H}_{t,:}\leftarrow
  \bm{H}_{t,:}+s\bm{z},
  \label{eq:inject}
\end{equation}
where $s$ is the intervention scale. The same intervention is applied during
prompt processing and at every autoregressive decoding step.

For each dataset and candidate block, we train its projector $P$ on a subset
$\mathcal{D}_{\mathrm{proj}}$ of the task's training split, keeping the
provided teacher vector and the student backbone frozen. Each example
comprises the prompt $x$ and the student-tokenized ground-truth final answer
$\bm{y}_S$ under teacher forcing. We fix $s=1$ during
training and minimize final-answer cross-entropy:
\begin{equation}
  \mathcal{L}_{\mathrm{proj}}
  =-\frac{1}{N_S}
  \sum_{t=1}^{N_S}
  \log p_S\!\left(y_{S,t}\mid x,
  y_{S,<t};\bm{z}\right),
  \label{eq:ceonly}
\end{equation}
where $y_{S,<t}$ is the ground-truth answer prefix, and conditioning on
$\bm{z}$ denotes its injection at the chosen block. Only $P$ is updated;
no intermediate solution tokens, teacher logits, or teacher hidden states
are needed, and no teacher--student hidden-state alignment is used in this
stage.

\subsection{Configuration Selection and Inference}
\label{sec:configuration}

We select the student injection block $L^\star$ and inference scale
$s^\star$ by accuracy on a fixed labeled support set drawn from the
official evaluation split. Each candidate block has a separately trained
projector and is evaluated one block at a time. We report accuracy on the
remaining evaluation examples, which are disjoint from the support set.
Support examples do not update the projectors or either backbone. Let
$P^\star$ denote the projector trained for the selected block. We then
precompute and store the intervention vector
\begin{equation}
  \bm{z}_{\mathrm{inf}}
  =s^\star P^\star\bm{v}_T.
  \label{eq:inference-vector}
\end{equation}
At inference, this vector is added to the output of student block
$L^\star$ at all prompt positions during prefill and at every decoding
step. Only the frozen student and the stored vector are needed; decoding
otherwise follows the baseline configuration.

\section{Experiments}
\label{sec:experiments}

\subsection{Experimental Setup}
\label{sec:setup}

\textbf{Teacher-side CoT Vector learning.} We learn a separate CoT Vector
for each dataset and candidate teacher block, using the dataset's entire
training split for two epochs. Each training example provides a prompt,
the dataset-provided solution trace, and a final answer. As described in
Section~\ref{sec:teacher-vectors}, the reference path processes the prompt
and solution trace including the final answer, while the injected path
processes the same prompt and
final answer with the vector applied at the candidate block. We optimize
the vector using the final-answer cross-entropy on the injected path and
representation-level alignment with the reference path
($\lambda_{\mathrm{CE}}=0.5$, $\tau=1$). The teacher backbone remains
frozen. Vector learning does not use the student model; once learned,
the vector is fixed and can be reused across student-side transfers.

\textbf{Student-side transfer.} We freeze the student backbone and the
teacher-side vector, and train only a bias-free linear projector using
final-answer cross-entropy (Eq.~\ref{eq:ceonly}). We shuffle the task
training split with a fixed seed, use its first $200$ examples to
train each projector for two epochs, and select the intervention
configuration on a disjoint support subset of the official evaluation
split. Student-side training uses the prompt and final answer,
without the dataset-provided gold solution trace. It requires the
fixed teacher-side vector but does not load or update the teacher
model. The projector is trained at injection scale $1$; the selected
scale is applied at inference.

\subsection{Math Reasoning Results}
\label{sec:main-results}

We evaluate three teacher--student settings from the Qwen2.5-Math
family~\citep{yang2024qwen25math}: the 7B-Instruct teacher with
the base 1.5B and 1.5B-Instruct students, and the 72B-Instruct
teacher with the base 7B student. We use GSM8K~\citep{cobbe2021gsm8k} and
MATH~\citep{hendrycks2021math}. The official training splits contain $7{,}473$ and $7{,}500$
examples, and the test splits contain $1{,}319$ and $5{,}000$
problems, respectively. For each candidate student block $L$, a projector maps the fixed
teacher-side vector to the student's residual space for injection at
block $L$. The projector dimensions are $1536\times3584$ for both 1.5B students
and $3584\times8192$ for the 7B student. For each math benchmark, we take $200$ approximately evenly spaced
problems from the official evaluation split as a fixed support set,
select the block and inference scale by support-set accuracy, and
report accuracy on the remaining $1{,}119$ GSM8K or $4{,}800$ MATH
problems. Neither the projector nor the backbones are trained on
support-set problems. The same support IDs are removed from every
baseline and injected-model evaluation. In Table~\ref{tab:math},
$\Delta$ is the injected-minus-student accuracy difference in
percentage points; Layer and Scale list the selected student block
and inference scale. Both baselines and injected models use
the Qwen2.5-Math chain-of-thought prompting format and zero-shot greedy
decoding with no input truncation and a maximum of $1{,}024$ and $2{,}048$
newly generated tokens for GSM8K and MATH, respectively; we report answer
accuracy using the Qwen2.5-Math extraction and grading protocol.

Table~\ref{tab:math} compares the three teacher--student pairs on
the held-out remainder of each official evaluation split. The two
base students gain $10.37$--$21.98$ percentage points across the four
settings; in particular, the 1.5B base student gains over $21$ points
on both GSM8K and MATH. Under the zero-shot CoT evaluation protocol,
the base students have relatively low baseline accuracy; the projected
intervention substantially improves their task performance, although
it does not close the gap to the teachers. The already strong
1.5B-Instruct student sees
little change ($-0.09$ points on GSM8K and $+0.34$ on MATH), possibly
because it has less room to benefit from the intervention.
These near-zero net changes conceal paired correctness transitions in
both directions (Appendix~\ref{sec:instruct-transitions},
Table~\ref{tab:instruct-transitions}).
Table~\ref{tab:math-support} in Appendix~\ref{sec:support-set}
reports the support-set scores used to select the student block and
inference scale; the $5.00$-point support-set gain for the
1.5B-Instruct student on GSM8K does not persist on the held-out
remainder. Table~\ref{tab:training-share} shows that the $200$ examples
used to train each math projector account for less than $3\%$ of the
corresponding official training split. The parameter counts and stored
intervention sizes appear in Table~\ref{tab:projection-footprint}.

\begin{table}[t]
\caption{Math reasoning accuracy (\%). $\Delta$ is Injected minus Student in percentage points; Layer is the zero-based student block; Scale is the inference scale.}
\label{tab:math}
\centering
\small
\setlength{\tabcolsep}{3pt}
\begin{tabular}{lllrrrrrr}
\toprule
Teacher model & Student model & Dataset & Student acc. & Teacher acc. & Injected & $\Delta$ & Layer & Scale \\
\midrule
7B-Instruct & 1.5B & GSM8K & 44.95 & 95.89 & 66.93 & +21.98 & 3 & 0.20 \\
7B-Instruct & 1.5B & MATH & 38.08 & 83.23 & 59.13 & +21.05 & 3 & 0.20 \\
72B-Instruct & 7B & GSM8K & 61.48 & 96.07 & 71.85 & +10.37 & 8 & 0.10 \\
72B-Instruct & 7B & MATH & 53.17 & 86.02 & 71.04 & +17.87 & 6 & 0.25 \\
7B-Instruct & 1.5B-Instruct & GSM8K & 86.33 & 95.89 & 86.24 & -0.09 & 5 & 0.10 \\
7B-Instruct & 1.5B-Instruct & MATH & 76.10 & 83.23 & 76.44 & +0.34 & 8 & 0.02 \\
\bottomrule
\end{tabular}
\end{table}

\begin{table}[t]
\caption{Projector-training data. Share is the percentage of the official training split used to train the projector.}
\label{tab:training-share}
\centering
\small
\begin{tabular}{lrrr}
\toprule
Dataset & Training split & Projector training & Share (\%) \\
\midrule
GSM8K & $7{,}473$ & $200$ & $2.68$ \\
MATH  & $7{,}500$ & $200$ & $2.67$ \\
\bottomrule
\end{tabular}
\end{table}

\begin{table}[t]
\caption{Student-side parameter and inference footprint. $|P|$ is the projector parameter count, $\bm{z}_{\mathrm{inf}}=s^\star P^\star\bm{v}_T$ is the stored intervention vector, and $|S|$ is the student parameter count.}
\label{tab:projection-footprint}
\centering
\small
\begin{tabular}{llrrr}
\toprule
Teacher model & Student model & $|P|$ & $|\bm{z}_{\mathrm{inf}}|$ & $|\bm{z}_{\mathrm{inf}}|/|S|$ (\%) \\
\midrule
7B-Instruct & 1.5B  & $5.51$M  & $1{,}536$ & $\approx 0.000102$ \\
72B-Instruct & 7B  & $29.36$M & $3{,}584$ & $\approx 0.000051$ \\
\bottomrule
\end{tabular}
\end{table}

\subsection{Comparison with Student-Side Training and Distillation}
\label{sec:baseline-comparison}

We compare projected CoT Vector transfer with LoRA~\citep{hu2022lora},
full-parameter supervised fine-tuning (SFT; \citealp{ouyang2022training}),
and DistiLLM~\citep{ko2024distillm} on MATH, using the
7B-Instruct-to-1.5B setting.
All adaptation methods start from the same Qwen2.5-Math-1.5B student checkpoint
and use the same $200$ training examples for two epochs of student-side
training. LoRA and SFT are trained on the dataset-provided full CoT
solutions, including the final answers; our student-side transfer uses
only final-answer supervision and updates only the projector. The
comparison matches the student-side data and epoch budgets, while
retaining each method's trainable components and supervision.

All methods use the same evaluation conditions: the Qwen2.5-Math CoT
prompting format, zero-shot greedy decoding with at most $2{,}048$
newly generated tokens, and the same answer-extraction and grading
protocol. We exclude the same $200$ support-set problems from every
method and report accuracy on the remaining $4{,}800$ MATH problems.

\begin{table}[t]
\caption{Comparison on MATH under matched student-side training-data and evaluation conditions. LoRA and SFT use full CoT supervision. Accuracy excludes the support set.}
\label{tab:math-baselines}
\centering
\small
\begin{tabular}{lr}
\toprule
Method & Accuracy (\%) \\
\midrule
Unadapted student & 38.08 \\
LoRA & 48.98 \\
SFT & 47.94 \\
DistiLLM & 49.20 \\
Projected CoT Vector (ours) & \textbf{59.13} \\
\bottomrule
\end{tabular}
\end{table}

Table~\ref{tab:math-baselines} shows that projected CoT Vector transfer
achieves $59.13\%$ accuracy, compared with $48.98\%$ for CoT-trained
LoRA, $47.94\%$ for CoT-trained SFT, and $49.20\%$ for DistiLLM.
With the same $200$ training examples and two student-side training
epochs, our method exceeds the strongest compared baseline, DistiLLM,
by $9.93$ percentage points. This advantage is obtained using only
final-answer supervision during student-side adaptation and keeping
the student backbone frozen, while the LoRA and SFT baselines receive
the full intermediate solution traces. Moreover, the trained projector
can be replaced at inference by a fixed $1{,}536$-dimensional vector,
providing a compact task-specific intervention that requires neither
the teacher nor backbone weight updates at deployment.

\subsection{Cross-Dataset Generalization}
\label{sec:cross-dataset}

We evaluate whether interventions learned for one math dataset also
transfer to other datasets. Using the 7B-Instruct teacher and the
base 1.5B student, we take the projectors trained on GSM8K or MATH and
the student blocks and inference scales selected on the corresponding
source-dataset support sets. We then evaluate each fixed configuration
on SVAMP~\citep{patel-etal-2021-nlp} and
ASDiv~\citep{miao-etal-2020-diverse}, without using either
target dataset to train the projector or select the intervention
configuration. Table~\ref{tab:cross-dataset} compares each transferred
configuration with the uninjected student on the same target dataset.

\begin{table}[t]
\caption{Cross-dataset evaluation on SVAMP and ASDiv (accuracy, \%). The source dataset determines projector training and support-set selection; accuracy and gain are measured on the target dataset.}
\label{tab:cross-dataset}
\centering
\small
\begin{tabular}{llrr}
\toprule
Target dataset & Source dataset & Accuracy (\%) & Gain (pp) \\
\midrule
SVAMP & No injection & 26.20 & --- \\
SVAMP & GSM8K & 70.10 & +43.90 \\
SVAMP & MATH & 62.30 & +36.10 \\
ASDiv & No injection & 33.90 & --- \\
ASDiv & GSM8K & 75.76 & +41.86 \\
ASDiv & MATH & 69.43 & +35.53 \\
\bottomrule
\end{tabular}
\end{table}

Both source-dataset configurations improve accuracy on each held-out
target dataset. On SVAMP, the GSM8K- and MATH-trained interventions
reach $70.10\%$ and $62.30\%$, respectively, versus a $26.20\%$
uninjected baseline; on ASDiv, they reach $75.76\%$ and $69.43\%$
versus $33.90\%$. The GSM8K-trained configuration yields the larger
gain on both target datasets.

\subsection{Code Generation Results}
\label{sec:code-results}

For code generation, we use Qwen2.5-32B-Instruct as the teacher
and Qwen2.5-7B-Instruct and the base Qwen2.5-7B as students. We evaluate
them on HumanEval~\citep{chen2021evaluating},
BigCodeBench~\citep{zhuo2024bigcodebench},
LiveCodeBench~\citep{jain2024livecodebench}, and
DS-1000~\citep{lai2023ds1000}.
We generate one completion
per problem with greedy decoding, no input truncation, and a configured
maximum number of newly generated tokens; we report
pass@1 using the corresponding execution tests. For each code benchmark, a support subset of its official evaluation
split is used to select the intervention configuration. We exclude the
same support problems from the injected and baseline pass@1 results
and report only the disjoint remainder. The Layer and Scale columns list the
selected student injection block (zero-based) and inference scale.
Baseline, Teacher, and Injected are evaluated on the same remaining
problem IDs. Table~\ref{tab:code-support} in
Appendix~\ref{sec:support-set} reports the support-set scores used for
configuration selection.

\begin{table}[t]
\caption{Code-generation pass@1 (\%). $\Delta$ is Injected minus Baseline in percentage points; Layer is the zero-based student block; Scale is the inference scale.}
\label{tab:code}
\centering
\small
\begin{tabular}{llrrrrrr}
\toprule
Dataset & Student & Baseline & Teacher & Injected & $\Delta$ & Layer & Scale \\
\midrule
HumanEval & 7B-Instruct & 72.81 & 78.07 & 77.19 & +4.38 & 7 & 0.05 \\
HumanEval & 7B & 55.26 & 78.07 & 64.91 & +9.65 & 27 & 0.05 \\
BigCodeBench & 7B-Instruct & 21.20 & 38.00 & 24.80 & +3.60 & 7 & 0.05 \\
BigCodeBench & 7B & 21.60 & 38.00 & 22.40 & +0.80 & 7 & 0.01 \\
LiveCodeBench & 7B-Instruct & 17.74 & 24.19 & 20.97 & +3.23 & 16 & 0.05 \\
LiveCodeBench & 7B & 6.45 & 24.19 & 6.45 & +0.00 & 20 & 0.05 \\
DS-1000 & 7B-Instruct & 50.00 & 57.33 & 54.67 & +4.67 & 12 & 0.01 \\
DS-1000 & 7B & 10.67 & 57.33 & 10.67 & +0.00 & 3 & 0.01 \\
\bottomrule
\end{tabular}
\end{table}

Table~\ref{tab:code} shows that the 7B-Instruct student also gains
$3.23$--$4.67$ pass@1 points across all four code-generation benchmarks,
which require executable programs rather than the numerical answers in
our math evaluation. The base 7B student gains $9.65$ points on
HumanEval and $0.80$ on BigCodeBench, but sees no change on
LiveCodeBench or DS-1000, where its low baseline pass@1 suggests that
a single-block intervention may be insufficient.

\subsection{Ablation: Direct Vector Training and Random-Vector Projection}
\label{sec:random-vector-controls}

We evaluate two controls on MATH and HumanEval to examine the
contribution of the teacher-side CoT Vector. The first directly trains
a vector in the student's residual space using final-answer
cross-entropy and injects it into the frozen student's residual stream,
without a projector. The second replaces the teacher-side CoT Vector
with a random vector in the teacher's residual space, rescaled to
match the CoT Vector's Euclidean norm, and trains a linear projector
to map it into the student's residual space. This control retains
projection-based adaptation while replacing the teacher-derived vector.

The MATH setting uses Qwen2.5-Math-7B-Instruct as the teacher and
the base Qwen2.5-Math-1.5B as the student; the HumanEval setting uses
Qwen2.5-32B-Instruct and Qwen2.5-7B-Instruct, respectively.
Table~\ref{tab:direct} compares both controls with the uninjected
student and the projected teacher CoT Vector. All scores exclude
the support-set examples, following the evaluation protocol above.

\begin{table}[t]
\caption{Direct student-vector training and random-vector projection ablations on MATH and HumanEval. For MATH, the teacher is Qwen2.5-Math-7B-Instruct and the student is Qwen2.5-Math-1.5B; for HumanEval, the teacher is Qwen2.5-32B-Instruct and the student is Qwen2.5-7B-Instruct. We report answer accuracy and pass@1 (\%), respectively, with support-set examples excluded.}
\label{tab:direct}
\centering
\small
\begin{tabular}{lrr}
\toprule
Intervention & MATH & HumanEval \\
\midrule
No injection & 38.08 & 72.81 \\
Direct student-vector training & 37.46 & 71.93 \\
Norm-matched random vector + projection & 42.63 & 70.18 \\
Teacher CoT Vector + projection & \textbf{59.13} & \textbf{77.19} \\
\bottomrule
\end{tabular}
\end{table}

On MATH, the teacher CoT Vector reaches $59.13\%$ accuracy, compared
with $37.46\%$ for the directly trained student vector and $42.63\%$ for
the norm-matched random vector with projection.
The corresponding HumanEval pass@1 scores are $77.19\%$, $71.93\%$,
and $70.18\%$, respectively, supporting the value of the teacher-derived
vector relative to direct student-vector training and norm-matched
random-vector projection.

\subsection{Case Study}
\label{sec:case-study}

Final-answer accuracy alone does not reveal how a model arrives at its
answer or whether it can distinguish valid reasoning from plausible
errors~\citep{wang2026can}; we therefore inspect a paired student
completion to illustrate a concrete change induced by the intervention.

We examine DS-1000 problem 351, which asks for the p-value of a two-tailed
t-test between two samples after removing NaN values. The zero-shot
Qwen2.5-7B-Instruct student identifies the appropriate
\lstinline{scipy.stats.ttest_ind} call but unpacks the return values in
the wrong order: \lstinline{p_value, _ = ...} assigns the t-statistic to
\lstinline{p_value}. It also does not assign the computed value to the
required output variable \lstinline{result}. This completion fails the
official execution tests.

\begin{lstlisting}[caption={Student without injection (fails).}]
import numpy as np
import scipy.stats
a, b = test_input
a_clean = a[~np.isnan(a)]
b_clean = b[~np.isnan(b)]
p_value, _ = scipy.stats.ttest_ind(a_clean, b_clean, equal_var=False)
\end{lstlisting}

With the selected intervention at student block $12$ and inference scale
$s{=}0.01$, the student instead assigns the second return value to
\lstinline{p_value} and writes \lstinline{result = p_value}. The
completion passes the official tests.

\begin{lstlisting}[caption={Student with injection (passes).}]
import numpy as np
import scipy.stats
a, b = test_input
a_clean = a[~np.isnan(a)]
b_clean = b[~np.isnan(b)]
_, p_value = scipy.stats.ttest_ind(a_clean, b_clean, equal_var=False)
result = p_value
\end{lstlisting}

This example shows a correction to a localized API-use and output-assignment
error while the student's overall solution strategy remains the same. It
illustrates one observed transition, not its frequency. Further code
examples and paired transition counts appear in
Appendix~\ref{sec:case-details}; math reasoning examples appear in
Appendix~\ref{sec:math-case-details}.

\section{Conclusion}
\label{sec:conclusion}

In this work, we studied projected cross-model CoT Vector transfer from a
strong teacher to a smaller student with a different residual dimension. Our
framework separates the transfer process into two stages. First, a
task-level CoT Vector is learned in the frozen teacher and then held fixed.
Second, a bias-free linear projector maps this vector into the student's
residual space while the student backbone remains frozen. Student-side
transfer trains only the projector on a small subset of the task training
data using final-answer
supervision; it does not require teacher logits, paired teacher--student
hidden states, or intermediate solution sequences. After selecting the
injection configuration, we precompute the intervention vector, so inference
requires only the frozen student and this fixed vector.

Across the evaluated tasks, we select intervention configurations on
labeled support subsets of the official evaluation splits and report
performance on their disjoint remainders. Across mathematical reasoning
and code generation, projected teacher vectors improve several smaller
frozen students despite differences in
model scale and residual dimension. Some settings show no gain or a
slight decrease, so the intervention is not uniformly beneficial.
Together with the small subset used to train each projector, these
results suggest that the teacher-side vector offers task-level guidance
that the student can use with limited answer supervision. Future
work should examine transfer across broader model families and task
domains, reduce reliance
on labeled support examples for configuration selection, and clarify
when an intervention helps or interferes with a student's existing
reasoning behavior.

\section*{AI Use Disclosure}

In this work, we used generative AI tools to aid and polish the writing, e.g., to improve the grammar, clarity, and readability of the manuscript. We have not used generative AI tools for other research tasks, including generating synthetic data, developing theoretical models or conceptual frameworks, formulating or proving mathematical claims, designing experiments, implementing methods, or interpreting results. All AI-assisted content was carefully reviewed and verified by all authors, and we take full responsibility for the final content of this work, including all text, claims, and artifacts.

\section*{Reproducibility Statement}

We make significant efforts to ensure the reproducibility of our results. Our method is implemented in PyTorch, and run with fixed random seeds. We will open-source the complete source code upon acceptance. All datasets used in this work are publicly available.

\bibliographystyle{iclr2027_conference}
\bibliography{iclr2027_conference}

\appendix
\section*{Appendix}


\section{Support-Set Results}
\label{sec:support-set}

The support set is used only to choose the student injection block and
inference scale. Tables~\ref{tab:math-support} and~\ref{tab:code-support} give the corresponding
accuracies; the main results exclude these examples.

\begin{table}[htbp]
\caption{Math support-set accuracy (\%). Layer is the zero-based student block; Scale is the inference scale.}
\label{tab:math-support}
\centering
\small
\setlength{\tabcolsep}{4pt}
\begin{tabular}{lllrrrr}
\toprule
Teacher model & Student model & Dataset & Student acc. & Injected & Layer & Scale \\
\midrule
7B-Instruct & 1.5B & GSM8K & 46.50 & 68.50 & 3 & 0.20 \\
7B-Instruct & 1.5B & MATH & 33.00 & 65.50 & 3 & 0.20 \\
72B-Instruct & 7B & GSM8K & 55.00 & 73.50 & 8 & 0.10 \\
72B-Instruct & 7B & MATH & 51.50 & 77.50 & 6 & 0.25 \\
7B-Instruct & 1.5B-Instruct & GSM8K & 79.00 & 84.00 & 5 & 0.10 \\
7B-Instruct & 1.5B-Instruct & MATH & 83.00 & 85.00 & 8 & 0.02 \\
\bottomrule
\end{tabular}
\end{table}

\begin{table}[htbp]
\caption{Code support-set pass@1 (\%). Layer is the zero-based student block; Scale is the inference scale.}
\label{tab:code-support}
\centering
\small
\setlength{\tabcolsep}{4pt}
\begin{tabular}{lllrrrr}
\toprule
Teacher model & Student model & Dataset & Student acc. & Injected & Layer & Scale \\
\midrule
32B-Instruct & 7B-Instruct & HumanEval & 82.00 & 86.00 & 7 & 0.05 \\
32B-Instruct & 7B & HumanEval & 76.00 & 84.00 & 27 & 0.05 \\
32B-Instruct & 7B-Instruct & BigCodeBench & 26.00 & 30.00 & 7 & 0.05 \\
32B-Instruct & 7B & BigCodeBench & 26.00 & 30.00 & 7 & 0.01 \\
32B-Instruct & 7B-Instruct & LiveCodeBench & 16.00 & 22.00 & 16 & 0.05 \\
32B-Instruct & 7B & LiveCodeBench & 6.00 & 6.00 & 20 & 0.05 \\
32B-Instruct & 7B-Instruct & DS-1000 & 64.00 & 70.00 & 12 & 0.01 \\
32B-Instruct & 7B & DS-1000 & 12.00 & 14.00 & 3 & 0.01 \\
\bottomrule
\end{tabular}
\end{table}


\section{Paired Correctness Transitions for the 1.5B-Instruct Student}
\label{sec:instruct-transitions}

Table~\ref{tab:instruct-transitions} compares the uninjected and injected
Qwen2.5-Math-1.5B-Instruct student for the 7B-Instruct teacher setting.
The counts use the same held-out problems and selected injection
configurations as Table~\ref{tab:math}; the support-set problems are
excluded. A transition records a change in answer correctness on the
same problem, not necessarily a change in the solution strategy.

\begin{table}[htbp]
\caption{Paired answer-correctness transitions on the held-out math sets.
Changed is the sum of fail-to-pass and pass-to-fail cases; Net is
fail-to-pass minus pass-to-fail.}
\label{tab:instruct-transitions}
\centering
\small
\begin{tabular}{lrrrrr}
\toprule
Dataset & Problems & Fail $\to$ pass & Pass $\to$ fail & Changed (\%) & Net \\
\midrule
GSM8K & $1{,}119$ & 43 & 44 & 87 (7.8) & $-1$ \\
MATH & $4{,}800$ & 183 & 167 & 350 (7.3) & $+16$ \\
\bottomrule
\end{tabular}
\end{table}

The transitions in opposite directions largely offset, leaving the
aggregate accuracy close to the uninjected baseline on both datasets.

\section{Math Case Study Details}
\label{sec:math-case-details}

Qualitative examples complement the quantitative results of
Table~\ref{tab:math} by showing how the intervention changes the student's
problem-solving behavior as reflected in the generated solutions. All
examples use the intervention reported in Table~\ref{tab:math} for the
7B-Instruct teacher with the 1.5B base student on GSM8K: a linear projector
at block $3$ with $s{=}0.20$. The zero-shot and injected outputs are
produced by the same student model with the same prompt and decoding
settings; correctness is determined by the official GSM8K answer
extraction~\citep{cobbe2021gsm8k}, and the examples are selected for
interpretability and are not used to compute additional metrics. Problems
are identified by their index in the official GSM8K test-set ordering. We
quote the model outputs verbatim, including any Python code they emit; this
code is not executed, and only the extracted final answer is judged.

\begin{table}[htbp]
\caption{Identifiers of the math case-study examples (all use the linear projector at block $3$ with $s{=}0.20$).}
\label{tab:math-case-details}
\centering
\small
\begin{tabular}{cllll}
\toprule
Case & Dataset & Problem ID & Baseline & With injection \\
\midrule
1 & GSM8K & 500 & fail & pass \\
2 & GSM8K & 387 & fail & pass \\
3 & GSM8K & 8 & fail & fail \\
4 & GSM8K & 4 & pass & fail \\
5 & GSM8K & 322 & fail & pass \\
\bottomrule
\end{tabular}

\vspace{2pt}
{\footnotesize In Case 5 the teacher also fails (10 boxes versus the ground-truth 9), so the injected student exceeds its teacher on this problem.}
\end{table}

\paragraph{Case 1 (GSM8K problem 500, fail $\to$ pass): repairing a
localized arithmetic error.}
The task asks how many insects Lily found, given a total of $43$ and two
linear relations among three people. The zero-shot student introduces the
variable $D$ for David, writes the correct equation
$D + (D + 7) + 2D = 43$, and then solves it incorrectly, claiming $D = 10$
and answering $17$. With vector injection, the student writes the same
equation with the same variable, expands the algebra step by step
($4D + 7 = 43$, $4D = 36$, $D = 9$), and answers $16$. The teacher solves the
problem using the same variable $D$.

The injected solution retains the same variable setup and equation structure
but corrects the algebraic manipulation, providing a concrete instance of
localized error correction without altering the student's overall solution
strategy.

\begin{lstlisting}[caption={Zero-shot student (fails).}]
Let's denote the number of insects David found as D. According to the problem, Lily found 7 more than David, so Lily found D + 7 insects. David found half of what Bodhi found, so Bodhi found 2D insects. Together, they found 43 insects, so we can write the equation: D + (D + 7) + 2D = 43. Solving this equation, we find that D = 10. Therefore, Lily found 10 + 7 = 17 insects. The answer is \boxed{17}.
\end{lstlisting}

\begin{lstlisting}[caption={Student with injection (passes).}]
Let's denote the number of insects David found as D. According to the problem, Lily found 7 more than David, so Lily found D + 7 insects. David found half of what Bodhi found, so Bodhi found 2D insects. The total number of insects collected is 43, so we can write the equation:

D + (D + 7) + 2D = 43

Combining like terms, we get:

4D + 7 = 43

Subtracting 7 from both sides, we get:

4D = 36

Dividing both sides by 4, we get:

D = 9

So, David found 9 insects. Since Lily found 7 more than David, Lily found 9 + 7 = 16 insects.
\end{lstlisting}

\paragraph{Case 2 (GSM8K problem 387, fail $\to$ pass): from a degenerate
output to a complete solution.}
The task asks for the total cost of six erasers and eight pencils. The
zero-shot student produces $5{,}593$ characters that echo a fragment of an
unrelated prompt template (``A box of chocolates costs\ldots'') and then
repeat the token \texttt{.Highlight} $502$ times. This token-level
repetition, a common failure mode of autoregressive generation, continues
until generation reaches the $1{,}024$-token GSM8K limit, and no final
answer can be extracted. With vector injection, the student produces a
$784$-character structured solution: it computes $6 \times 2 = 12$ and
$8 \times 3 = 24$, sums them, verifies the arithmetic with a short Python
program, and returns the correct answer $36$.

This example shows the intervention restoring a complete, structured solution
from a degenerate output that never attempted the problem.

\begin{lstlisting}[caption={Zero-shot student (fails; excerpt, 5,593 characters in total).}]
A box of chocolates costs $10. How much do 10 boxes of chocolates cost?.Highlight
.Highlight
.Highlight
... (the token .Highlight repeats 502 times in total)
\end{lstlisting}

\begin{lstlisting}[caption={Student with injection (passes).}]
Let's break down the problem step by step and then use Python to verify the solution.

1. Calculate the cost of 6 erasers.
2. Calculate the cost of 8 pencils.
3. Add the two amounts to get the total cost.

The cost of 6 erasers is \(6 \times 2 = 12\) dollars.
The cost of 8 pencils is \(8 \times 3 = 24\) dollars.
The total cost is \(12 + 24 = 36\) dollars.

Now, let's use Python to verify this calculation.
```python
# Define the cost of erasers and pencils
cost_eraser = 2
cost_pencil = 3

# Calculate the total cost for 6 erasers and 8 pencils
total_cost_erasers = 6 * cost_eraser
total_cost_pencils = 8 * cost_pencil
total_cost = total_cost_erasers + total_cost_pencils

print(total_cost)
```
```output
36
```
The total cost for 6 erasers and 8 pencils is \(\boxed{36}\) dollars.
\end{lstlisting}

\paragraph{Case 3 (GSM8K problem 8, the transfer boundary): byte-identical
outputs.}
The task is a multi-step distance problem. The zero-shot student repeats the
problem statement verbatim and then repeats the token \texttt{oud} $460$
times. With vector injection, its complete generated string is byte-identical
to the zero-shot output, including whitespace and line breaks ($2{,}706$
characters in total), and therefore also fails.

This example illustrates the limits of the intervention: when the model does
not enter a problem-solving state (i.e., it never produces a structured
solution), the injected representation produces no change in the output.

\begin{lstlisting}[caption={Zero-shot student and student with injection (byte-identical, fails; excerpt, 2,706 characters in total).}]
John drives for 3 hours at a speed of 60 mph and then turns around because he realizes he forgot something very important at home. He tries to get home in 4 hours but spends the first 2 hours in standstill traffic. He spends the next half-hour driving at a speed of 30mph, before being able to drive the remaining time of the 4 hours going at 80 mph. How far is he from home at the end of those 4 hours?
 oud
 oud
 oud
... (the token oud repeats 460 times in total)
\end{lstlisting}

\paragraph{Case 4 (GSM8K problem 4, a failure introduced by the
intervention).}
The task asks how many cups of feed remain for the final meal of a flock of
$20$ chickens. The zero-shot student answers correctly. The flock eats
$3 \times 20 = 60$ cups per day; since $15 + 25 = 40$ cups have already been
given, $20$ cups remain for the final meal. With vector injection, the
student performs the same computation but then divides the remaining $20$
cups by the flock size and answers $1$ cup per chicken, interpreting the
question as ``how much feed does each chicken need in the final meal''
rather than ``how many cups of feed are needed in total''. This is a
semantic misalignment error: the computation is correct for a different
question, but wrong for the actual task.

Unlike the first two examples, in which the intervention repairs the
student's solution, here it shifts the student's interpretation of the
question and turns a correct answer into a failure.

\begin{lstlisting}[caption={Zero-shot student (passes; excerpt, Python verification code omitted).}]
1. Wendi gives her chickens 15 cups of feed in the morning and 25 cups of feed in the afternoon. So, the total amount of feed given in the morning and afternoon is \(15 + 25 = 40\) cups.
2. Since Wendi gives her chickens a total of 3 cups of feed per chicken per meal, and she has 20 chickens, the total amount of feed she gives her chickens in a day is \(3 \times 20 = 60\) cups.
3. Therefore, the amount of feed she needs for the final meal is \(60 - 40 = 20\) cups.
... (Python verification code omitted)
The amount of feed Wendi needs to give her chickens in the final meal of the day is \(\boxed{20}\) cups.
\end{lstlisting}

\begin{lstlisting}[caption={Student with injection (fails; excerpt, Python verification code omitted).}]
1. Calculate the total amount of feed given to the chickens in the morning and afternoon:
   Total feed given = 15 cups + 25 cups = 40 cups
2. Calculate the remaining amount of feed needed for the day:
   Remaining feed = 60 cups - 40 cups = 20 cups
3. Calculate the amount of feed each chicken needs in the final meal:
   Feed per chicken in the final meal = 20 cups / 20 chickens = 1 cup per chicken
So, Wendi needs to give each chicken 1 cup of feed in the final meal.
... (Python verification code omitted)
The amount of feed each chicken needs in the final meal is \(\boxed{1}\) cup.
\end{lstlisting}

\paragraph{Case 5 (GSM8K problem 322, exceeding the teacher).}
The task asks how many boxes are needed to store $26 + 46$ pencils, given
that three boxes plus two loose pencils contain $26$ pencils. The teacher
solves $3x + 2 = 26$ correctly to get $x = 8$ pencils per box. However, it
rounds the box count up separately for each person ($\lceil 26/8 \rceil = 4$
and $\lceil 46/8 \rceil = 6$) and concludes $4 + 6 = 10$ boxes, missing the
optimal packing: combining the pencils first gives $72/8 = 9$ boxes. The
official ground-truth answer is $9$ boxes; the teacher's $10$ boxes satisfy
the storage constraint but are not minimal, and the benchmark marks them
incorrect. The zero-shot student degenerates into template spam
(\lstinline|((\boxed{?}) + (\boxed{?}) = ...)|). With vector injection, the
student combines the pencils first, divides by $8$, verifies the result with
a short Python program, and returns $9$, which the teacher does not produce.

This example shows that the transferred vector can encode problem-solving
structure that the teacher does not apply consistently, and that the injected
student can exceed the teacher on individual problems.

\begin{lstlisting}[caption={Teacher (fails; excerpt, the omitted middle follows the same per-person rounding).}]
Let's denote the number of pencils in each box by \( x \). According to the problem, we can write the following equation:

\[ 3x + 2 = 26 \]

...
\[ x = 8 \]

So, each box contains 8 pencils.
... (the teacher rounds 26/8 = 3.25 up to 4 boxes for Jam and 46/8 = 5.75 up to 6 boxes for Meg)
To find the total number of boxes needed for both Jam and Meg, we add the number of boxes each needs:

\[ 4 + 6 = 10 \]

Therefore, Jam and Meg need \(\boxed{10}\) boxes to store all their pencils.
\end{lstlisting}

\begin{lstlisting}[caption={Student with injection (passes).}]
Let's break down the problem step by step.

1. First, we need to determine how many pencils Jam has in total. We know that Jam has three boxes of pencils and 2 loose pencils, and the total number of pencils is 26. Let's denote the number of pencils in each box as \( x \). So, the total number of pencils Jam has can be expressed as:
\[ 3x + 2 = 26 \]
2. We can solve this equation to find the value of \( x \):
\[ 3x + 2 = 26 \]
\[ 3x = 24 \]
\[ x = 8 \]
So, each box contains 8 pencils.
3. Next, we need to find out how many boxes Jam and Meg need to store all their pencils. We know that Jam has 26 pencils and Meg has 46 pencils. The total number of pencils is:
\[ 26 + 46 = 72 \]
4. Since each box contains 8 pencils, the total number of boxes needed is:
\[ \frac{72}{8} = 9 \]
So, Jam and Meg need 9 boxes to store all their pencils.

Let's write the Python code to verify this solution.
```python
# Number of pencils in each box
pencils_per_box = (26 - 2) // 3

# Total number of pencils Jam and Meg have
total_pencils = 26 + 46

# Total number of boxes needed
total_boxes = total_pencils // pencils_per_box

print(total_boxes)
```
```output
9
```
The total number of boxes needed to store all the pencils is \(\boxed{9}\).
\end{lstlisting}

Together, the examples illustrate the range of behaviors behind the aggregate
gains. The intervention can:
\begin{enumerate}[label=(\arabic*)]
\item repair localized computation errors while preserving solution
structure (problem 500);
\item restore structured solutions from degenerate outputs (problems 387
and 322);
\item leave degenerate outputs unchanged when the model does not enter a
solving state (problem 8); and
\item occasionally introduce semantic misalignment errors (problem 4).
\end{enumerate}
These behaviors are consistent with the quantitative results and the design
hypothesis: the intervention is most effective when the student already
possesses the basic problem-solving capability needed to make use of the
injected representation.
\section{Code-Generation Case Study Details}
\label{sec:case-details}

The transition counts below concern only problems for which paired
zero-shot and injected outputs were available: $190$ problems in
DS-1000 and $262$ in BigCodeBench. They therefore have different
denominators from the main evaluation scores in Table~\ref{tab:code}.
For these paired problems, DS-1000 has $14$ fail-to-pass and $2$
pass-to-fail transitions; BigCodeBench has $28$ and $15$, respectively.
Problems without paired outputs are not included in these counts.

Qualitative examples complement the quantitative results of
Table~\ref{tab:code} by showing how the intervention changes the student's
behavior as reflected in the generated programs. The zero-shot and injected
outputs are produced by the same student model with the same prompt and
decoding settings; correctness is determined by the corresponding official
execution tests. The examples are selected for interpretability and are not
used to compute additional metrics.

\begin{table}[htbp]
\caption{Identifiers of the code case-study examples.}
\label{tab:code-case-details}
\centering
\small
\begin{tabular}{clllll}
\toprule
Case & Dataset & Problem ID & Student & Baseline & With injection \\
\midrule
1 & DS-1000 & 351 & 7B-Instruct & fail & pass \\
2 & DS-1000 & 486 & 7B-Instruct & fail & pass \\
2 & DS-1000 & 486 & 7B (base) & fail & fail \\
\bottomrule
\end{tabular}
\end{table}

\paragraph{Case 1 (DS-1000 problem 351, fail $\to$ pass): correcting the
returned value and output assignment.} The task asks for the p-value of a
two-tailed t-test between two samples after removing NaN values. The
zero-shot student correctly identifies and calls
\lstinline{scipy.stats.ttest_ind}, but makes two completion errors. First,
it unpacks the return values in the wrong order:
\lstinline{p_value, _ = ...} assigns the t-statistic to
\lstinline{p_value}. Second, it never assigns the computed value to the
expected output variable. With vector injection, the student unpacks the
return values correctly,
\lstinline{_, p_value = ...}, and assigns
\lstinline{result = p_value}. The resulting program passes the official
tests using the linear projector at block $12$ and $s{=}0.01$.

This example provides a concrete instance of the intervention correcting
a localized API-use and output-formatting error while preserving the
student's overall solution strategy.

\begin{lstlisting}[caption={Zero-shot student (fails).}]
import numpy as np
import scipy.stats
a, b = test_input
a_clean = a[~np.isnan(a)]
b_clean = b[~np.isnan(b)]
p_value, _ = scipy.stats.ttest_ind(a_clean, b_clean, equal_var=False)
\end{lstlisting}

\begin{lstlisting}[caption={Student with injection (passes).}]
import numpy as np
import scipy.stats
a, b = test_input
a_clean = a[~np.isnan(a)]
b_clean = b[~np.isnan(b)]
_, p_value = scipy.stats.ttest_ind(a_clean, b_clean, equal_var=False)
result = p_value
\end{lstlisting}

\paragraph{Case 2 (DS-1000 problem 486, the transfer boundary): contrasting
two students on the same problem.} The task asks the model to normalize a
two-dimensional NumPy array globally. Because
\lstinline{MinMaxScaler} normalizes each feature column by default, a
valid solution must reshape the array into a single column, apply the
scaler, and reshape the result back. The import of
\lstinline{MinMaxScaler} is provided by the benchmark's code template and
is omitted from the listings below.

Both students are evaluated with the same prompt, decoding procedure, and
injection configuration: a linear projector at block $3$ with
$s{=}0.01$. This is the base student's best configuration on DS-1000 and
does not favor the instruct student: at this configuration, the
instruct student scores $52.0$, below its best score over the grid. Thus,
the comparison does not attribute the base student's failure to a
configuration selected specifically for the instruct student.

The zero-shot instruct student identifies the appropriate reshaping
strategy but fails to instantiate the scaler, resulting in a
\lstinline{NameError}. With vector injection, it produces a
self-contained one-line solution and passes the tests. In contrast, the
zero-shot base student does not produce an executable solution in this
example. Instead, it repeats the problem statement and the evaluation
template, including the harness's placeholder instructions. Its complete generated string with
injection is byte-identical to the zero-shot output, including whitespace
and line breaks (1,980 characters in total), and therefore also fails.

This example mirrors the aggregate results in Table~\ref{tab:code}. For
the instruct student, the intervention repairs a localized completion
error while preserving an otherwise appropriate strategy. For the base
student, the same intervention produces no textual change because the
model does not produce an executable solution in this example. Within
this example and the broader results, this behavior is consistent with
the interpretation that the transfer is more effective when the student
already possesses the basic instruction-following and problem-solving
capabilities needed to make use of the injected representation.

\begin{lstlisting}[caption={Zero-shot instruct student (fails).}]
result_one_column = scaler.fit_transform(a.reshape(-1, 1))
result = result_one_column.reshape(a.shape)
\end{lstlisting}

\begin{lstlisting}[caption={Instruct student with injection (passes).}]
result = MinMaxScaler().fit_transform(a.reshape(-1, 1)).reshape(a.shape)
\end{lstlisting}

\begin{lstlisting}[caption={Base student, zero-shot and with injection
(byte-identical, fails; excerpt, 1,980 characters in total).}]
Problem:
Right now, I have my data in a 2D numpy array `a`. If I was to use
MinMaxScaler fit_transform on the array, it will normalize it column by
column, whereas I wish to normalize the entire np array all together. ...
result = ... # put solution in this variable
BEGIN SOLUTION
<code>
Below is a code template with a [insert] placeholder. Complete the code by
outputting ONLY the code that should replace [insert]. ...
\end{lstlisting}

Together, the two cases illustrate the complementary aspects of the
transfer boundary in Table~\ref{tab:code}: the intervention repairs a
localized completion error for the instruct student, while the same
intervention leaves the base student's non-executable output unchanged.

\end{document}